\def\year{2021}\relax
\documentclass[letterpaper]{article} % DO NOT CHANGE THIS
\usepackage{aaai21}  % DO NOT CHANGE THIS
\usepackage{times}  % DO NOT CHANGE THIS
\usepackage{helvet} % DO NOT CHANGE THIS
\usepackage{courier}  % DO NOT CHANGE THIS
\usepackage[hyphens]{url}  % DO NOT CHANGE THIS
\usepackage{graphicx} % DO NOT CHANGE THIS
\usepackage{natbib}  % DO NOT CHANGE THIS AND DO NOT ADD ANY OPTIONS TO IT
\usepackage{caption} % DO NOT CHANGE THIS AND DO NOT ADD ANY OPTIONS TO IT
\usepackage{tikz}
\usetikzlibrary{graphs}
\usepackage{amsfonts}
\usepackage{amsmath}
\usepackage{caption}
\usepackage{subcaption}
\usepackage{array}
\usepackage{threeparttable}
\usepackage{multirow}
\usepackage{booktabs}
\usepackage{times}
\usepackage{textcomp}

\usepackage{makecell}

\title{Train a One-Million-Way Instance Classifier for Unsupervised Visual \\ Representation Learning}
\author{Yu Liu, Lianghua Huang, Pan Pan, Bin Wang, Yinghui Xu, Rong Jin\\}
\affiliations{
    Machine Intelligence Technology Lab, Alibaba Group\\
    \{ly103369, xuangen.hlh, panpan.pp, ganfu.wb, renji.xyh, jinrong.jr\}@alibaba-inc.com
}
\begin{document}

\maketitle

\begin{abstract}
    This paper presents a simple unsupervised visual representation learning method with a pretext task of discriminating all images in a dataset using a parametric, instance-level classifier.
    The overall framework is a replica of a supervised classification model, where \textit{semantic classes} (e.g., \textit{dog, bird,} and \textit{ship}) are replaced by \textit{instance IDs}.
    However, scaling up the classification task from thousands of \textit{semantic labels} to millions of \textit{instance labels} brings specific challenges including 1) the large-scale softmax computation; 2) the slow convergence due to the infrequent visiting of instance samples; and 3) the massive number of negative classes that can be noisy.
    This work presents several novel techniques to handle these difficulties.
    First, we introduce a hybrid parallel training framework to make large-scale training feasible.
    Second, we present a raw-feature initialization mechanism for classification weights, which we assume offers a contrastive prior for instance discrimination and can clearly speed up converge in our experiments.
    Finally, we propose to smooth the labels of a few hardest classes to avoid optimizing over very similar negative pairs.
    While being conceptually simple, our framework achieves competitive or superior performance compared to state-of-the-art unsupervised approaches, i.e., SimCLR, MoCoV2, and PIC under ImageNet linear evaluation protocol and on several downstream visual tasks, verifying that full instance classification is a strong pretraining technique for many semantic visual tasks.
\end{abstract}

\noindent Unsupervised visual representation learning has recently shown encouraging progress~\cite{moco2020,simclr2020}.
Methods using \textit{instance discrimination} as a pretext task~\cite{cmc2019,moco2020,simclr2020} have demonstrated competitive or even superior performance compared to supervised counterparts under ImageNet~\cite{imagenet2009} linear evaluation protocol and on many downstream visual tasks.
This shows the potential of unsupervised representation learning methods since they can utilize almost unlimited data without manual labels.

\begin{figure}[t]
\centering
  \includegraphics[width=0.45\textwidth]{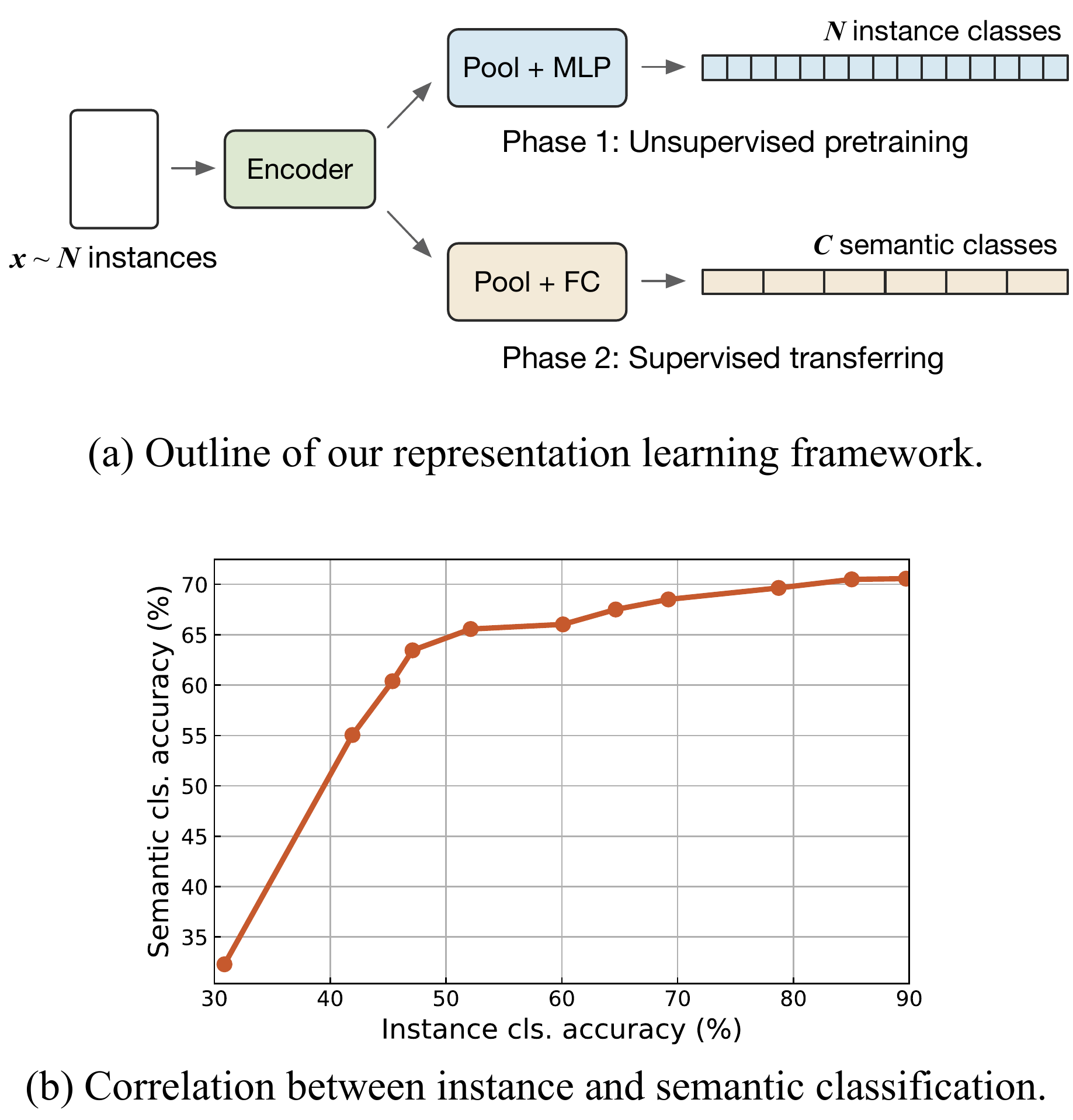}
  \caption{
    (a) An overview of our unsupervised visual representation learning framework.
    Without manual labels, we simply train an instance-level classifier that tries to \textit{distinguish all images in a dataset} to learn discriminative representations that can be well transferred to supervised tasks.
    (b) The relationship between instance (unsupervised) and semantic (supervised) classification accuracies.
    We observe a strong positive correlation between them in our experiments.  % TODO: this sentence is TOO weak
  }
\label{fig:1}
\end{figure}

To solve the \textit{instance discrimination} task, usually a dual-branch structure is used, where two transformed views of a same image are encouraged to get close, while transformed views from different images are expected to get far apart~\cite{moco2020,simclr2020,mocov2_2020}.
These methods often rely on specialized designs such as memory bank~\cite{instdisc2018}, momentum encoder~\cite{moco2020,mocov2_2020}, large batch size~\cite{simclr2020,simclrv2_2020}, or shuffled batch normalization (BN)~\cite{moco2020,mocov2_2020} to compensate for the lack of negative samples or handle the information leakage issue (i.e., samples on a same GPU tend to get closer due to shared BN statistics).

\begin{figure*}[t]
% TODO: draw backward pass; improve the caption; match the caption
\centering
\includegraphics[width=0.85\textwidth]{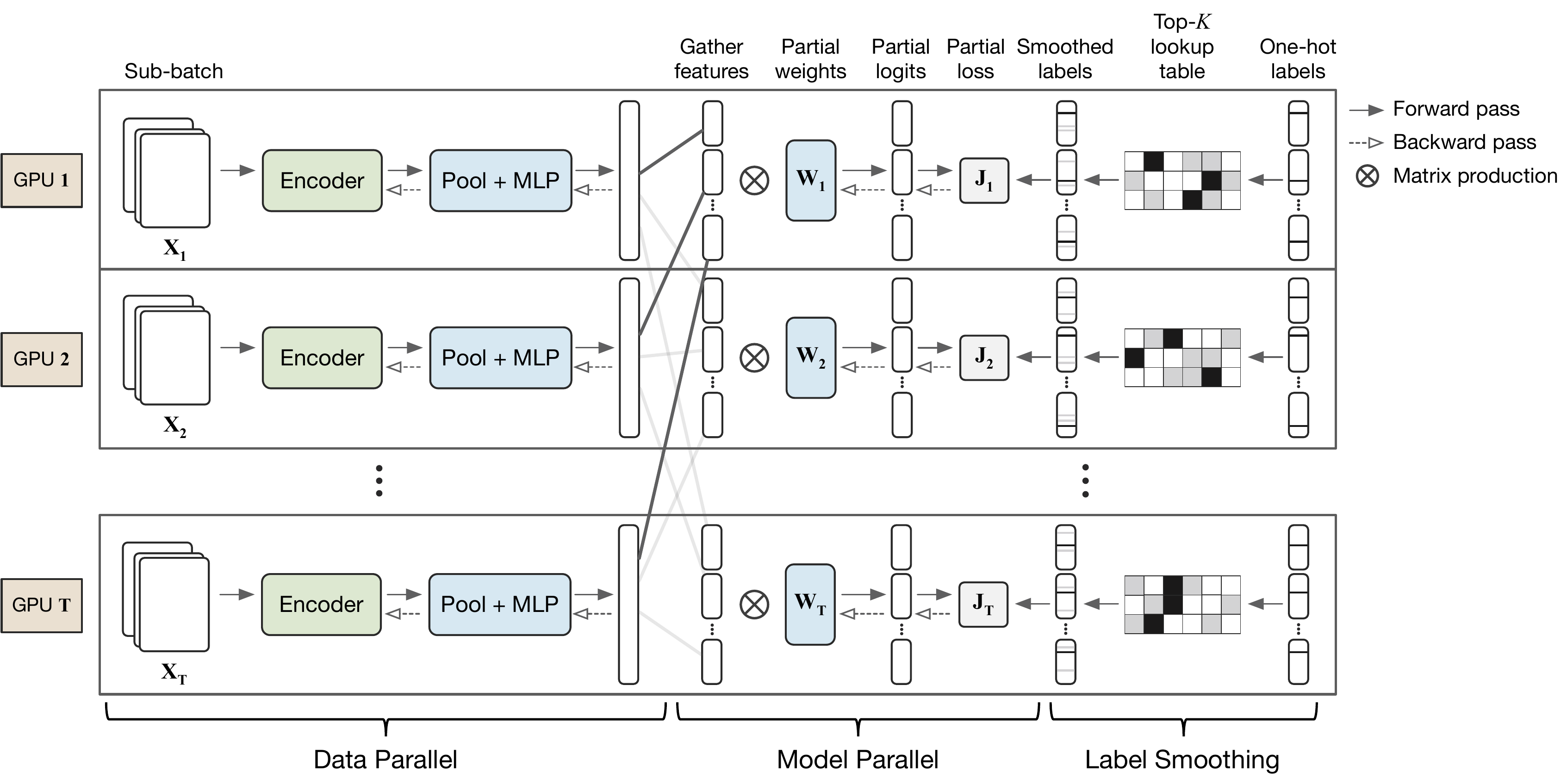}
\caption{
    An outline of our distributed hybrid parallel (DHP) training process on $T$ GPU nodes.
    \textit{Data parallel:} Following \textit{data parallel} mechanism, we copy the encoding and MLP layers to all nodes, each processing a subset of minibatch data.
    \textit{Model parallel:} Following \textit{model parallel} mechanism, we evenly divide the classification weights to different nodes, and distribute the computation of classification scores (forward pass) and weight/feature gradients (backward pass) to different GPUs.
    % On each node,, we first gather features from all other nodes and compute partial logits using local classification weights.
    % Then we sum exponential values of logits over all nodes to compute the softmax probabilities and the partial loss.
    % Finally, we compute the gradients of the loss to features and weights, sum the gradients over all nodes, and run a step of optimization.
    % We repeat the process to iteratively optimize for better representation.
    \textit{Label smoothing:} We smooth labels of the top-$K$ hardest negative classes for each instance to avoid optimizing over noisy pairs.
    % We find the top-$K$ hardest negative classes for each instance, and smooth their labels from $y_- = 0$ to $y_- = \alpha / K$ to avoid optimizing over very similar negative pairs.
}
\label{fig:2}
\end{figure*}

Unlike dual-branch approaches, one-branch scheme (e.g., parametric instance-level classification) usually avoids the information leakage issue, and can potentially explore a larger set of negative samples.
ExemplarCNN~\cite{discriminative2014} and PIC~\cite{parametric2020} are of this category.
Nevertheless, due to the high GPU computation and memory overhead of large-scale instance-level classification, these methods are either tested on small datasets~\cite{discriminative2014} or rely on negative class sampling~\cite{parametric2020} to make training feasible.

This work summarizes typical challenges of using one-branch instance discrimination for unsupervised representation learning, including 1) the large-scale classification, 2) the slow convergence due to the infrequent instance access, and 3) a large number of negative classes that can be noisy, and proposes novel techniques to handle them.
First, we introduce a hybrid parallel training framework to make large-scale classifier training feasible.
It relies on model parallelism that divides classification weights to different GPUs, and evenly distribute the softmax computation (both in forward and backward passes) to them.
Figure~\ref{fig:2} shows an overview of our distributed training process.
This training framework can theoretically support up to 100-million-way classification using 256 GPUs~\cite{song2020large}, which far exceeds the number of 1.28 million ImageNet-1K instances, indicating the scalability of our method.

Second, instance classification faces the slow convergence problem due to the extremely infrequent visiting of instance samples~\cite{parametric2020}.
In this work, we tackle this problem by introducing a \textit{a contrastive prior} to the instance classifier.
% In this work, we present a raw-feature initialization mechanism for the classifier to speed up convergence.
Specifically, with a randomly initialized network, we fix all but BN layers of it and run an inference epoch to extract all instance features;
then we directly assign these features to classification weights as an initialization.
The intuition is two-fold.
% On the one hand, assigning instance features $\mathbf{x}_i$ to classification weights $\mathbf{w}_i$ makes $\mathbf{w}_i$ a closer proxy of instance $i$, and potentially increases the visiting frequency of $\mathbf{x}_i$ in early epochs.
% On the one hand, initializing classification weights as instance features in essence converts the classification task to a pair-wise instance comparison task, providing a warm start for convergence.
On the one hand, we presume that \textit{running BNs} may offer a contrastive prior in the output instance features, since in each iteration the features will subtract a weighted average of other instance features extracted in previous iterations.
On the other hand, initializing classification weights as instance features in essence converts the classification task to a pair-wise instance comparison task, providing a warm start for convergence.
% On the one hand, assigning instance features $\mathbf{x}_i$ to classification weights $\mathbf{w}_i$ makes them closer proxies for instances, especially in early epochs.
% On the one hand, assigning instance features $\mathbf{x}_i$ to classification weights $\mathbf{w}_i$ significantly increases the visiting frequency of instances' \textit{close proxies} $\mathbf{w}_i$, especially in early epochs.
% On the other hand, we presume that \textit{running BNs} offer a contrastive prior in the output instance features, since in each iteration the features will subtract a weighted average of other instance features extracted in previous iterations.
% The assumption is also experimentally validated in our ablation study.

Finally, regarding the massive number of negative instance classes that significantly raises the risk of optimizing over very similar negative pairs, we propose to smooth the labels of the top-$K$ hardest negative classes to make training easier.
Specifically, we compute cosine similarities between instance proxies -- their corresponding classification weights -- and find the negative classes with the top-$K$ highest similarities for each instance.
The labels of these classes are smoothed by a factor of $\alpha$ (i.e., from $y_- = 0$ to $y_- = \alpha / K$).
The right part of Figure~\ref{fig:2} shows the smoothing process.
Note these top-$K$ indices are computed once per training epoch, which is very efficient and adds only minimal computational overhead for the training process.

We evaluate our method under the common ImageNet linear evaluation protocol as well as on several downstream tasks related to detection or fine-grained classification.
Despite its simplicity, our method shows competitive results on these tasks.
For example, our method achieves a top-1 accuracy of 71.4\% under ImageNet linear classification protocol, outperforming all other instance discrimination based methods~\cite{simclr2020,mocov2_2020,parametric2020}.
We also obtain a semi-supervised accuracy of 81.8\% on ImageNet-1K when only 1\% of labels are provided, surpassing previous best result by around 4.7\%.
We hope our full instance classification framework can serve as a simple and strong baseline for the unsupervised representation learning community.

\section{Related Work}

\noindent \textbf{Unsupervised visual representation learning.}
Unsupervised or self-supervised visual representation learning aims to learn discriminative representation from visual data where no manual labels are available.
Usually a pretext task is utilized to determine the quality of the learned representation and to iteratively optimize the parameters.
Representative pretext tasks include transformation prediction~\cite{transforms2018gidaris,transforms2019zhang}, in-painting~\cite{inpainting2016}, spatial or temporal patch order prediction~\cite{order2015doersch,order2016unoroozi,order2018noroozi}, colorization~\cite{colorization2016}, clustering~\cite{clustering2018caron,clustering2019zhuang,clustering2020}, data generation~\cite{generation2018jenni,bigbigan2019,generation2016donahue}, geometry~\cite{geometry2015}, and a combination of multiple pretext tasks~\cite{combine2017doersch,combine2019feng}.

\noindent \textbf{Contrastive visual representation learning.}
More recently, contrastive representation learning methods~\cite{cpcv22019,moco2020} have shown significant performance improvements by using strong data augmentation and proper loss functions~\cite{mocov2_2020,simclr2020}.
For these methods, usually a dual-branch structure is employed, where two augmented views of an image are encouraged to get close while augmented views from different images are forced to get far apart.
One problem of these methods is the shortage of negative samples.
Some methods rely on large batch size~\cite{simclr2020}, memory bank~\cite{instdisc2018}, or momentum encoder~\cite{moco2020,mocov2_2020} to enlarge the negative pool.
Another issue is regarding the information leakage issue~\cite{moco2020,mocov2_2020} where features extracted on a same GPU tend to get close due to the shared BN statistics.
MoCo~\cite{moco2020,mocov2_2020} solves this problem by using shuffled batch normalization (BN), while SimCLR~\cite{simclr2020} handles the problem with a synchronized global BN.

\noindent \textbf{Instance discrimination for representation learning.}
Unlike the two-branch structure used in contrastive methods, some approaches~\cite{discriminative2014,parametric2020} employ a parametric, one-branch structure for instance discrimination,
which avoids the information leakage issue.
Exemplar-CNN~\cite{discriminative2014} learns a classifier to discriminate between a set of ``surrogate classes'', each class represents different transformed patches of a single image.
Nevertheless, it shows worse performance than non-parametric approaches~\cite{instdisc2018}.
PIC~\cite{parametric2020} improves Exemplar-CNN in two ways:
1) it introduces a sliding-window data scheduler to alleviate the infrequent instance visiting problem;
2) it utilizes recent classes sampling to reduce the GPU memory consumption.
Despite its effectiveness, it relies on complicated scheduling and optimization processes, and it cannot fully explore the large number of negative instances.
This work presents a much simpler instance discrimination method that uses an ordinary data scheduler and optimization process.
In addition, it is able to make full usage of the massive number of negative instances in every training iteration.

\section{Methodology}

\subsection{Overall Framework}

This work presents an unsupervised representation learning method with a pretext task of classifying all image instances in a dataset.
Figure~\ref{fig:1} (a) shows the outline of our method.
% The pipeline of the method is similar to common supervised classification that consists of 5 components:
% in each training iteration, we
% 1) randomly sample a minibatch of images from the dataset,
% 2) perform data augmentation on minibatch samples,
% 3) extract representation vectors of these images using an encoder network,
% 4) map the representations to embeddings where a linear instance-level classifier is applied, and
% 5) compute the softmax classification loss and run a step of model optimization.
% We repeat the process to iterate for better representation.
The pipeline is similar to common supervised classification, where \textit{semantic classes} are replaced by \textit{instance IDs}.
Inspired by the design improvements used in recent unsupervised frameworks~\cite{simclr2020}, we slightly modify some components, including using stronger data augmentation (i.e., random crop, color jitter, and Gaussian blur), a two-layer MLP projection head, and a cosine softmax loss.
The cosine softmax loss is defined as
\begin{eqnarray}
\label{eq:cross_entropy}
J = -\frac{1}{|I|} \sum_{i\in I} \log \frac{\exp(\cos(\mathbf{w}_i, \mathbf{x}_i) / \tau)}{\sum_{j=1}^N \exp(\cos(\mathbf{w}_j, \mathbf{x}_i) / \tau)},
\end{eqnarray}
where $I$ denotes the indices of sampled image instances in a minibatch, $\mathbf{x}_i$ is the projected embedding of instance $i$, $\mathbf{W} = \{\mathbf{w}_1, \mathbf{w}_2, \cdots, \mathbf{w}_N\} \in \mathbb{R}^{D\times N}$ represents the instance classification weights, $\cos(\mathbf{w}_j, \mathbf{x}_i) = (\mathbf{w}_j^{\mathrm{T}} \mathbf{x}_i) / (\|\mathbf{w}_j\|_2 \cdot \|\mathbf{x}_i\|_2)$ denotes the cosine similarity between $\mathbf{w}_j$ and $\mathbf{x}_i$, and $\tau$ is a temperature adjusting the scale of cosine similarities.

Nevertheless, there are still challenges for this vanilla instance classification model to learn good representation, including
1) the large-scale instance classes (e.g., 1.28 million instance classes for ImageNet-1K dataset);
2) the extremely infrequent visiting of instance samples; and
3) the massive number of negative classes that makes training difficult.
We propose three efficient techniques to improve the representation learning and the scalability of our method:
\begin{itemize}
\item \textit{Hybrid parallelism.} To support large-scale instance classification, we rely on hybrid parallelism and evenly distribute the softmax computation (both in forward and backward passes) to different GPUs. Figure~\ref{fig:2} shows a schematic of the distributed training process on $T$ GPUs.
\item \textit{A contrastive prior.} To improve the convergence, we propose to introduce a \textit{a contrastive prior} to the instance classifier. This is simply achieved by initializing classification weights as raw instance features extracted by a \textit{fixed} random network with \textit{running BNs}.
% To improve the convergence, we propose to initialize the classification weights as raw instance features extracted by a random initial network, where all but BN layers of it are fixed. We assume that such initialization offers a contrastive prior for instance discrimination.
\item \textit{Smoothing labels of hardest classes.} the massive number of negative classes raises the risk of optimizing over very similar pairs. We apply label smoothing on the top-$K$ hardest instance classes to alleviate this issue.
\end{itemize}

Note that the above improvements only bring little or no computational overhead for the training process.
Next we will introduce these techniques respectively in details.

\begin{figure}[t]
\centering
  \includegraphics[width=0.36\textwidth]{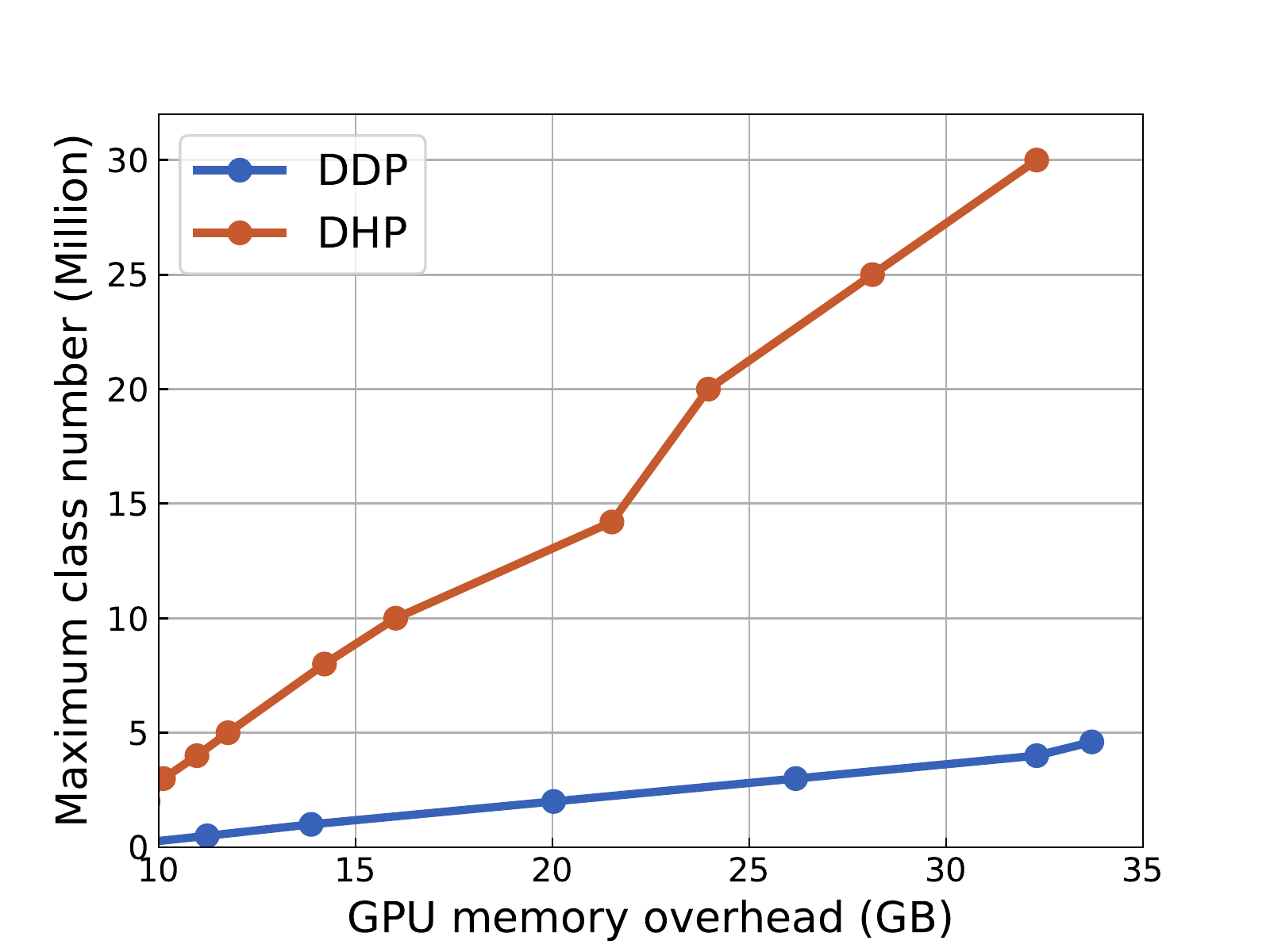}
  \caption{
      Comparison of the maximum number of classes supported by DDP and DHP training frameworks under different GPU memory constraints.
  }
\label{fig:3}
\end{figure}

\begin{table}[t]
\small
\begin{center}
\caption{
    Comparison of the GPU memory consumption and the training time per epoch of DDP and DHP training frameworks.
    Experiments are conducted on ImageNet-1K and ImageNet-21K datasets.
}
\label{tab:gpu_memory}
\begin{threeparttable}
\begin{tabular}{
    >{\raggedright\arraybackslash} m{1.9cm}
    >{\centering\arraybackslash} m{1.15cm}
    >{\centering\arraybackslash} m{0.75cm}
    >{\centering\arraybackslash} m{0.6cm}
    >{\centering\arraybackslash} m{0.9cm}
    >{\centering\arraybackslash} m{0.75cm}}
    \Xhline{2.8\arrayrulewidth}
    \multirow{2}{4em}{Dataset} & \multirow{2}{5em}{\#Instances} & \multicolumn{2}{c}{DDP} & \multicolumn{2}{c}{DHP} \\
    & & Memory & Time & Memory & Time \\
    \hline
    ImageNet-1K & 1.28M & 15.6GB & 428s & \textbf{9.18GB} & \textbf{302s} \\
    ImageNet-21K & 14.2M & \textit{OOM}\tnote{*} & - & \textbf{21.51GB} & \textbf{2940s} \\
    \Xhline{2.8\arrayrulewidth}
\end{tabular}
\begin{tablenotes}
    \item[*] \textit{OOM}: Out-of-memory.
\end{tablenotes}
\end{threeparttable}
\end{center}
\end{table}

\subsection{Hybrid Parallelism}

Training an instance classifier usually requires to learn a large-scale \textit{fc} layer.
For example, for ImageNet-1K with approximately 1.28 million images, one needs to optimize a weight matrix of size $\mathbf{W}\in \mathbb{R}^{D\times 1280000}$.
For ImageNet-21K, the size further enlarged to $\mathbf{W}\in \mathbb{R}^{D\times 14200000}$.
This is often infeasible when using a regular distributed data parallel (DDP) training pipeline.
In this work, we introduce a distributed hybrid parallel (DHP) training framework~\cite{song2020large} to make large-scale classification feasible.

Figure~\ref{fig:2} summarizes the outline of the distributed hybrid parallel training process on $T$ GPU nodes.
For encoding and MLP layers, we follow the \textit{data parallel} pipeline and \textbf{copy} them to different GPUs, each processing a subset of minibatch data;
while for the large-scale \textit{fc} layer, we follow the \textit{model parallel} mechanism and \textbf{split} the weights evenly to $T$ GPUs.
At each training iteration and for each GPU node, we
1) extract features of a subset of minibatch samples;
2) gather features from all other nodes;
3) compute partial cosine logits using local classification weights;
4) compute the exponential values of logits and sum them over all nodes to obtain the softmax denominators;
5) compute softmax probabilities and the cross-entropy loss on the subset data;
6) deduce gradients of the local loss with respect to features and weights;
7) gather feature gradients from all GPU node and sum them;
8) run a step of optimization to update parameters of encoding, MLP, and classification layers.
The pipeline is repeated to loop through the complete dataset for several epochs to optimize for better representation.

Figure~\ref{fig:3} compares the GPU memory overhead of DDP and DHP training frameworks when increasing the (pseudo) class number from 10K to 30M.
The experiment is conducted on 64 V100 GPUs with 32GB memory and a total batch size of 4096.
DDP reports out-of-memory (OOM) error when the class number reaches 4.7 million, while the DHP training framework can support up to 30 million number of classes, which is $\mathbf{6.4\times}$ of the DDP's limit.
We also note that the DHP can benefit from more GPUs to support larger-scale instance classification, but DDP does not bear this scalability.
Table~\ref{tab:gpu_memory} compares the training efficiency of DDP and DHP frameworks on ImageNet-1K and ImageNet-21K under the same batch size settings.
We show that the DHP training framework not only consumes less GPU memory, but also trains much faster than the DDP counterpart.

% \begin{table}[t]
% \small
% \begin{center}
% \caption{Ablation study on the effectiveness of different components in our method.}
% \label{tab:init_methods}
% \begin{threeparttable}
% \begin{tabular}{
%     >{\raggedright\arraybackslash} m{3.0cm}
%     >{\centering\arraybackslash} m{1.6cm}}
%     \Xhline{2.8\arrayrulewidth}
%     Init. method & Linear eval. (epoch 10) \\
%     \hline
%     Random init. & 12.4 \\
%     Raw with fixed BNs & 22.7 \\
%     Raw with running BNs & \textbf{27.4} \\
%     \Xhline{2.8\arrayrulewidth}
% \end{tabular}
% \end{threeparttable}
% \end{center}
% \end{table}

\begin{figure}[t]
\centering
  \includegraphics[width=0.40\textwidth]{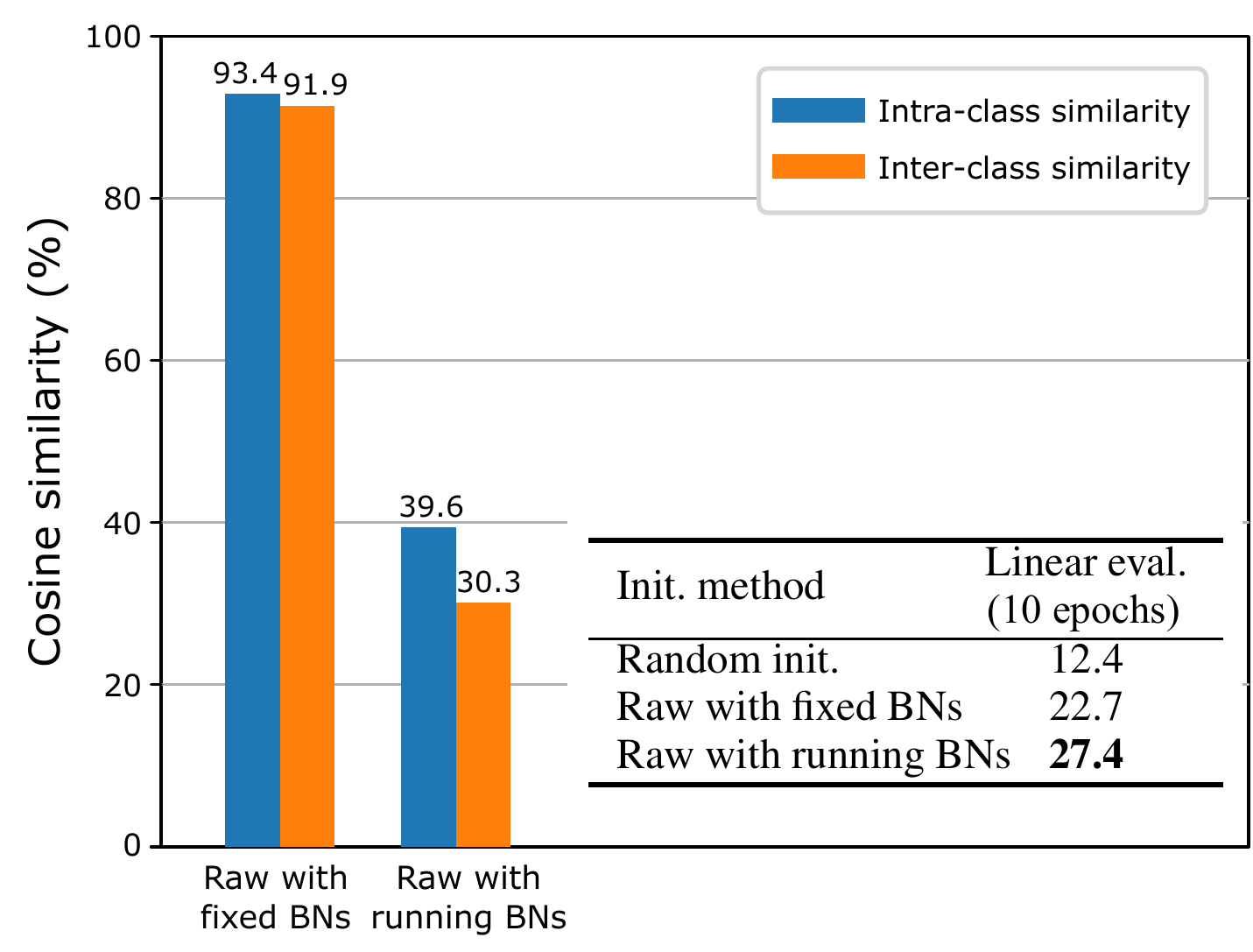}
  \caption{
      \textit{Table:} comparison of ImageNet linear evaluation accuracies of different weight-initialization methods, evaluated after $10$ epochs training.
      \textit{Bar chart:} comparison of average intra- and inter-instance-class cosine similarities when using different initialization methods.
  }
\label{fig:4}
\end{figure}

\subsection{A Contrastive Prior}\label{sec:initialization}

Instance classification faces the slow convergence problem in early epochs due to the infrequent visiting of instance samples (i.e., once access per epoch).
A recent work~\cite{parametric2020} handles this infrequent visiting problem by using a sliding-window data scheduler, which samples overlapped batches between adjacent iterations.
This increases the positive instance visiting but it also significantly multiplies the time of looping over the complete dataset.

In this work, we handle this problem from a different perspective: we propose to speed up the convergence by introducing a contrastive prior to classification weights.
% In this work, we present a simple weight initialization mechanism to handle the convergence issue.
Specifically, before training started, we run an inference epoch using the \textit{fixed} random initial network with \textit{running BNs} to extract all instance features $\mathbf{X} = \{\mathbf{x}_1, \mathbf{x}_2, \cdots, \mathbf{x}_N\} \in \mathbb{R}^{D\times N}$;
then we directly assign them to classification weights $\mathbf{W} = \{\mathbf{w}_1, \mathbf{w}_2, \cdots, \mathbf{w}_N\} \in \mathbb{R}^{D\times N}$ as an initialization.
The intuition behind this initialization mechanism is two-fold.
First, running BNs can offer a contrastive prior in the output features, since in each inference phase, the features computed after every BN layer will subtract a running average of other instance features extracted in previous iterations.
Second, assigning features to weights approximately converts the classification task to a pair-wise metric learning task in early epochs, which is relatively easier to converge and offers a warm start for instance classification.

% First, assigning instance features to weights approximately converts the classification task to a pair-wise metric learning task in early epochs, which is relatively easier to converge and provides a warm start for instance classification.
% First, assigning instance features to weights converts the parametric classification task to a pair-wise instance comparison task, providing a warm start for convergence.
% First, a weight vector $\mathbf{w}_i$ represents the feature center of different transformed views of instance $i$.
% Initializing $\mathbf{w}_i$ 
% First, a weight vector $\mathbf{w}_i$ represents the feature center of different transformed views of instance $i$, thus initializing weights as instance features can significantly increase the access of instances' \textit{close proxies} in early epochs.
% Second, running BNs can offer a contrastive prior in the output features, since in each inference phase, the features extracted after every BN layer will subtract a running average of other instance features extracted in all previous iterations.

Figure~\ref{fig:4} compares the discriminative ability of different classifier initialization schemes, i.e., random weight initialization (\textit{random init.} for short), raw-feature initialization with fixed BNs (\textit{raw with fixed BNs} for short), and raw-feature initialization with running BNs (\textit{raw with running BNs} for short).
We use ImageNet linear evaluation accuracy (evaluated after $10$ epochs training) as well as average intra- and inter-instance-class similarities as the indicators.
As shown in Figure~\ref{fig:4}, \textit{raw with running BNs} achieves the best linear evaluation accuracy, and clearly outperforms other initialization methods.
In addition, \textit{raw with running BNs} obtains a much larger similarity gap ($\sim9.3\%$) between positive and negative instance pairs than \textit{raw with fixed BNs} ($\sim1.5\%$),
validating the assumption that running BNs may provide a contrastive prior for instance discrimination.
% The results validate our assumption that running BNs may provide a contrastive prior for instance discrimination.

We also note that the instance features extracted by a random network with running BNs are also a robust start for semantic classification.
We run an instance retrieval experiment on the \textit{train} set of ImageNet-1K with a randomly initialized ResNet-50 network to extract all image features, and determine whether the searched instance and the query instance are of a same semantic category.
We find that a 3\% top-1 accuracy can be achieved, which far exceeds the top-1 accuracy of 0.1\% of a random guess.

% \begin{figure*}[t]
% \centering
% \includegraphics[width=0.80\textwidth]{figure5}
% \caption{Visualization of i2i instance retrieval.}
% \label{fig:5}
% \end{figure*}

\begin{table}[t]
\small
\begin{center}
\caption{Ablation study on the effectiveness of different components in our method.}
\label{tab:ablation_components}
\begin{threeparttable}
\begin{tabular}{
    >{\centering\arraybackslash} m{1.0cm}
    >{\centering\arraybackslash} m{1.8cm}
    >{\centering\arraybackslash} m{1.4cm} |
    >{\centering\arraybackslash} m{0.9cm}
    >{\centering\arraybackslash} m{0.9cm}}
    \Xhline{2.8\arrayrulewidth}
    \multirow{2}{4em}{MLP head} & \multirow{2}{5em}{A contrastive prior} & \multirow{2}{4em}{Label smoothing} & \multicolumn{2}{c}{ImageNet} \\
    &  &  & Top-1 & Top-5 \\
    \hline
    & \checkmark & \checkmark & 58.6 & 83.1 \\
    \checkmark & & & 67.3 & 87.7 \\
    \checkmark & \checkmark & & 67.6 & 88.0 \\
    \checkmark & \checkmark & \checkmark & \textbf{68.2} & \textbf{88.5} \\
    \Xhline{2.8\arrayrulewidth}
\end{tabular}
\end{threeparttable}
\end{center}
\end{table}

\begin{table}[t]
\small
\begin{center}
\caption{Ablation study that compares full instance classification and sampled instance classification.}
\label{tab:ablation_instance_sampling}
\begin{threeparttable}
\begin{tabular}{
    >{\centering\arraybackslash} m{2.6cm} |
    >{\centering\arraybackslash} m{1.2cm}
    >{\centering\arraybackslash} m{1.2cm}}
    \Xhline{2.8\arrayrulewidth}
    \#Sampled instances & Top-1 & Top-5 \\
    \hline
    \\[-0.9em]
    $2^{10}$ & 64.8 & 86.3 \\[0.05em]
    $2^{12}$ & 65.3 & 86.7 \\[0.05em]
    $2^{14}$ & 65.4 & 86.7 \\[0.05em]
    $2^{16}$ & 65.5 & 86.8 \\[0.05em]
    \textbf{Full} & \textbf{67.3} & \textbf{87.7} \\
    \Xhline{2.8\arrayrulewidth}
\end{tabular}
\end{threeparttable}
\end{center}
\end{table}

\begin{table}[t]
\small
\begin{center}
\caption{Ablation study that compares Gaussian random and a contrastive prior for classifier initialization.}
\label{tab:ablation_initialization}
\begin{threeparttable}
\begin{tabular}{
    >{\centering\arraybackslash} m{1.2cm} |
    >{\centering\arraybackslash} m{2.0cm}
    >{\centering\arraybackslash} m{2.0cm}}
    \Xhline{2.8\arrayrulewidth}
    \#Epochs & Gaussian random & A contrastive prior \\
    \hline
    10 & 12.4 & 27.4 (\textbf{+15.0}) \\
    25 & 40.8 & 46.3 (\textbf{+5.5}) \\
    50 & 56.1 & 58.0 (\textbf{+1.9}) \\
    100 & 62.9 & 64.1 (\textbf{+1.2}) \\
    200 & 67.3 & 67.6 (\textbf{+0.3}) \\
    400 & 69.3 & 69.7 (\textbf{+0.4}) \\
    \Xhline{2.8\arrayrulewidth}
\end{tabular}
\end{threeparttable}
\end{center}
\end{table}

\begin{table}[t]
\small
\begin{center}
\caption{Ablation study of label smoothing with different number of hard classes $K$ and smoothing factor $\alpha$.}
\label{tab:label_smoothing}
\begin{threeparttable}
\begin{tabular}{
    >{\centering\arraybackslash} m{1.5cm}
    >{\centering\arraybackslash} m{1.4cm} |
    >{\centering\arraybackslash} m{1.0cm}
    >{\centering\arraybackslash} m{1.0cm}}
    \Xhline{2.8\arrayrulewidth}
    Hard class number $K$ & Smoothing factor $\alpha$ & Top-1 & Top-5 \\
    \hline
    & & & \\[-0.95em]
    \multicolumn{2}{c}{\textit{no smoothing}} & 67.6 & 88.0 \\[0.05em]
    100 & 0.1 & 67.7 & 88.2 \\
    100 & 0.2 & \textbf{68.2} & \textbf{88.5} \\
    100 & 0.3 & 67.5 & 88.2 \\
    50 & 0.2 & 68.0 & 88.5 \\
    200 & 0.2 & 67.9 & 88.4 \\
    \Xhline{2.8\arrayrulewidth}
\end{tabular}
\end{threeparttable}
\end{center}
\end{table}

\subsection{Smoothing Labels of Hardest Classes}

A challenge of instance-level classification is that it introduces a very large number of negative classes, significantly raising the risk of optimizing over very similar pairs that can be noisy and make the training hard to converge.

In this work, we handle this problem by applying label smoothing on a few hardest instance classes.
Although other techniques (e.g., clustering) are also applicable, we choose label smoothing for its simplicity and efficiency.
We notice that the semantically similar instance pairs are relatively stable across the training process.
Therefore, we represent each instance $i$ as its corresponding classification weights $\mathbf{w}_i$ (instead of its unstable features $\mathbf{x}_i$), and compute the cosine similarities between $\mathbf{w}_i$ and all other weights $\mathbf{W}_{\bar{i}} = \{\mathbf{w}_1, \cdots, \mathbf{w}_{i-1}, \mathbf{w}_{i+1}, \cdots, \mathbf{w}_N\} \in \mathbb{R}^{D\times (N - 1)}$ to find the top-$K$ hardest negative classes $H_-^i = \{c_1, c_2, \cdots, c_K\}$.
The label of class $j \in \{1, 2, \cdots, N\}$ is then defined as
\begin{eqnarray}
y_j^i = \left\{
\begin{aligned}
    1 - \alpha,&~~~~j = i,\\
    \alpha / K,&~~~~j \in H_-^i,\\
    0,&~~~~\text{otherwise}.
\end{aligned}
\right.
\end{eqnarray}
The loss function in Eq.~\eqref{eq:cross_entropy} is redefined as
\begin{eqnarray}
\label{eq:smooth_cross_entropy}
J = -\frac{1}{|I|} \sum_{i\in I} \log \frac{\sum_{j=1}^N y_j^i \exp(\cos(\mathbf{w}_j, \mathbf{x}_i) / \tau)}{\sum_{j=1}^N \exp(\cos(\mathbf{w}_j, \mathbf{x}_i) / \tau)}.
\end{eqnarray}
The top-$K$ similarities between instance weights are computed once per epoch, which only amounts for a small fraction of training time.
The smoothed softmax cross-entropy loss reduces the impact of noisy or very similar negative pairs on the learned representation.
This is also verified later in our ablation study, where smoothing labels of several hardest classes improves the transfer performance.

\begin{table}[t]
\small
\begin{center}
\caption{State-of-the-art comparison of linear classification accuracy of unsupervised methods on ImageNet-1K.}
\label{tab:linear_evaluation}
\begin{threeparttable}
\begin{tabular}{
    >{\raggedright\arraybackslash} m{5.85cm}
    >{\centering\arraybackslash} m{0.8cm}
    >{\centering\arraybackslash} m{0.8cm}}
    \Xhline{2.8\arrayrulewidth}
    \textbf{Method} & \textbf{Top-1} & \textbf{Top-5} \\
    \hline
    \textit{Supervised} & 76.3 & 93.1 \\
    Exemplar~\cite{discriminative2014} & 48.6 & - \\
    CPC~\cite{cpc2018} & 48.7 & 73.6 \\
    InstDisc.~\cite{instdisc2018} & 54.0 & - \\
    BigBiGAN~\cite{bigbigan2019} & 56.0 & 77.4 \\
    NPID++~\cite{npidpp2018} & 59.0 & - \\
    LocalAgg.~\cite{localagg2019} & 60.2 & - \\
    MoCo~\cite{moco2020} & 60.6 & - \\
    SeLa~\cite{sela2020} & 61.5 & 84.0 \\
    PIRL~\cite{pirl2020} & 63.6 & - \\
    CPCv2~\cite{cpcv22019} & 63.8 & 85.3 \\
    CMC~\cite{cmc2019} & 64.1 & 85.4 \\
    SimCLR~\cite{simclr2020} & 69.3 & 89.0 \\
    PIC~\cite{parametric2020} & 70.8 & 90.0 \\
    MoCoV2~\cite{mocov2_2020} & 71.1 & - \\
    \textbf{Ours} & \textbf{71.4} & \textbf{90.3} \\
    \Xhline{2.8\arrayrulewidth}
\end{tabular}
\end{threeparttable}
\end{center}
\end{table}

\section{Experiments}

\subsection{Experiment Settings}
% The training and evaluation datasets as well as the implementation details of our unsupervised method are summarized in the following:

\noindent \textbf{Training datasets.}
Unless specified, we use ImageNet-1K to train our unsupervised model for most experiments.
ImageNet-1K consists of around 1.28 million images belonging to 1000 classes.
We treat every image instance (along with its various transformed views) in the dataset as a unique class, and train a 1.28 million-way instance classifier as a pretext task to learn visual representation.

\noindent \textbf{Evaluation datasets.}
The learned visual representations are evaluated in three ways:
First, under the linear evaluation protocol of ImageNet-1K, we fix the representation model and learn a linear classifier upon it.
The \textit{top-1/top-5} classification accuracies are employed to compare different unsupervised methods.
Second, we evaluate the semi-supervised learning performance on ImageNet-1K, where methods are required to classify images in the \textit{val} set when only a small fraction (i.e., 1\% or 10\%) of manual labels are provided in the \textit{train} set.
Third, we evaluate the transferring performance by finetuning the representations on several downstream tasks and compute performance gains.
In our experiments, downstream tasks include Pascal-VOC object detection~\cite{pascalvoc2010}, iNaturalist18 fine-grained image classification~\cite{inaturalist2018}, and many others.

\noindent \textbf{Implementation details.}
We use ResNet-50~\cite{resnet2016} as the backbone in all our experiments.
We train our model using the SGD optimizer, where the weight decay and momentum are set to 0.0001 and 0.9, respectively.
The initial learning rate (\textit{lr}) is set to 0.48 and decays using the cosine annealing scheduler.
In addition, we use 10 epochs of linear \textit{lr} warmup to stabilize training.
The minibatch size is 4096 and the feature dimension $D=128$.
% We train on 64 V100 32GB GPUs with a minibatch size of 4096.
% The initial learning rate is set to 0.48 and decay after each training epoch using the cosine annealing scheduler.
% In addition, we use 10 epochs of linear warmup for the learning rate to stabilize training.
% The initial learning rate is set to 0.48 and decayed using cosine annealing scheduler. A 1-epoch linear warmup is adopted for 10 training epochs, 3-epoch warmup for less than 100 training epochs, and 10-epoch warmup for others.
We set the temperature in Eq.~\eqref{eq:cross_entropy} as $\tau = 0.15$, and the smoothing factor in Eq.~\eqref{eq:smooth_cross_entropy} as $\alpha = 0.2$.
For fair comparison, following practices in recent works~\cite{simclr2020,parametric2020}, we feed two augmented views per instance for training.
All experiments are conducted on 64 V100 GPUs with 32GB memory.

% We use the same data augmentations as used in SimCLR.
% For fair comparison, similar to MoCo, SimCLR, and PIC
%  unless specified. The same data augmentations as~\cite{simclr2020} are adopted, and for fair comparison, two data-augmented views are fed for training, although not compulsory for our method.

\begin{table}[t]
\small
\begin{center}
\caption{Comparison of semi-supervised learning accuracy of both \textit{label propagation} and \textit{representation learning} based methods on ImageNet-1K.}
\label{tab:semi_supervised}
\begin{threeparttable}
\begin{tabular}{
    >{\raggedright\arraybackslash} m{5.8cm}
    >{\centering\arraybackslash} m{0.7cm}
    >{\centering\arraybackslash} m{0.7cm}}
    \Xhline{2.8\arrayrulewidth}
    \multirow{2}{4em}{\textbf{Method}} & \multicolumn{2}{c}{\textbf{Label Fraction}} \\
    & \textbf{1\%} & \textbf{10\%} \\
    \hline
    \textit{Supervised} & 48.4 & 80.4 \\
    \\[-0.9em]
    \textit{Label propagation:} & & \\[0.2em]
    ~~PseudoLabels~\cite{pseudolabels2019} & 51.6 & 82.4 \\
    ~~VAT+Entropy Min.~\cite{vat2018} & 47.0 & 83.4 \\
    ~~UDA~\cite{uda2019} & - & 88.5 \\
    ~~FixMatch~\cite{fixmatch2020} & - & 89.1 \\
    \\[-0.9em]
    \textit{Representation Learning:} & & \\[0.2em]
    ~~InstDisc.~\cite{discriminative2014} & 39.2 & 77.4 \\
    ~~PIRL~\cite{pirl2020} & 57.2 & 83.8 \\
    ~~PCL~\cite{pcl2020} & 75.6 & 86.2 \\
    ~~SimCLR~\cite{simclr2020} & 75.5 & 87.8 \\
    ~~PIC~\cite{parametric2020} & 77.1 & 88.7 \\
    ~~\textbf{Ours} & \textbf{81.8} & \textbf{89.2} \\
    \Xhline{2.8\arrayrulewidth}
\end{tabular}
\end{threeparttable}
\end{center}
\end{table}

\subsection{Ablation Study}
\label{sec:ablation_study}

This section validates several modeling and configuration options for our method.
We compare the quality of representations using ImageNet linear protocol evaluated on the \textit{val} set.
In each experiment, the linear classifier is trained with a batch size of 2048 and a \textit{lr} of 40 that decays during training under the cosine annealing rule.

% This section validates several modeling and configuration options for our method.
% We compare the quality of learned representations using ImageNet linear evaluation protocol.
% Under this protocol, we freeze the unsupervised pretrained backbone and learn a supervised linear classifier above the normalized output features on the \textit{train} set of ImageNet. The linear classifier is trained for 100 epochs with batch size of 2048, initial learning rate of 40 and cosine learning rate scheduler, weight decay of 0.
% The top-1 and/or top-5 accuracies on the \textit{val} set are used as indicators.

\noindent \textbf{Ablation: effectiveness of components.}
Table~\ref{tab:ablation_components} shows the linear evaluation performance using different combinations of components in our method, including a two-layer MLP head, a contrastive prior, and label smoothing.
Accuracies are measured after 200-epochs training.
We find that all the three components bring performance gains, improving the top-1 accuracy of our method from 58.6\% to a competitive 68.2\%.
We also observe that a vanilla instance classification model can already achieve a top-1 accuracy of 67.3\%, suggesting that full instance classification is a very strong baseline for unsupervised representation learning.
The contrastive prior and label smoothing on the top-$K$ hardest classes further boost the linear evaluation accuracy by around 1\%.

\noindent \textbf{Ablation: full instance classification \textit{v.s.} sampled instance classification.}
Table~\ref{tab:ablation_instance_sampling} compares linear classification accuracies using representations learned by full instance classification and by sampled instance classification, with sampling sizes ranging from $2^{10}$ to $2^{16}$.
Note we remove \textit{a contrastive prior} and \textit{label smoothing} in the experiments and only analyze the impact of class sampling.
We observe that full instance classification clearly outperforms sampled instance classification by a margin of 1.8\%, verifying the benefits of exploring the complete set of negative instances.

\noindent \textbf{Ablation: a contrastive prior \textit{v.s.} random initialization.}
Table~\ref{tab:ablation_initialization} compares the linear evaluation performance of our method using Gaussian random and a contrastive prior for classifier initialization, with training length increased from 10 to 400 epochs.
% Table~\ref{tab:ablation_initialization} compares linear evaluation performance (top-1 accuracy) of our method using random and raw-BN-instance-feature initialized classification weights, with training length increased from 10 to 400 epochs.
We observe that a contrastive prior significantly speeds up convergence compared to random initialization, especially in early epochs (i.e., epoch 10, 25, and 50).
Besides, the accuracy of a contrastive prior version consistently outperforms random initialization counterpart, showing the robustness of our initialization mechanism.

\noindent \textbf{Ablation: label smoothing on hardest classes.}
Table~\ref{tab:label_smoothing} shows the impact of label smoothing on representation learning.
We vary the considered number $K$ of hardest negative classes, and the smoothing factor $\alpha$.
A \textit{no smoothing} baseline where $K = 0$ and $\alpha = 0$ is also included for comparison.
We observe that smoothing labels of a few hardest classes improves linear evaluation performance over non-smoothing baseline in most hyper-parameter settings.
The best accuracy can be obtained with $K = 100$ and $\alpha = 0.2$, where a 0.6\% gain of top-1 accuracy can be achieved.

% \noindent \textbf{Ablation: qualitative analysis.}
%

\begin{table}[t]
\small
\begin{center}
\caption{Comparison of transferring performance on PASCAL VOC object detection.}
\label{tab:pascal_voc}
\begin{threeparttable}
\begin{tabular}{
    >{\raggedright\arraybackslash} m{3.7cm}
    >{\centering\arraybackslash} m{0.7cm}
    >{\centering\arraybackslash} m{0.7cm}
    >{\centering\arraybackslash} m{0.7cm}}
    \Xhline{2.8\arrayrulewidth}
    \textbf{Method} & \textbf{AP} & \textbf{AP50} & \textbf{AP75} \\
    \hline
    \textit{Supervised} & 53.5 & 81.3 & 58.8 \\
    MoCo~\cite{moco2020} & 55.9 & 81.5 & 62.6 \\
    MoCoV2~\cite{mocov2_2020} & \textbf{57.4} & \textbf{82.5} & \textbf{64.0} \\
    PIC~\cite{parametric2020} & 57.1 & 82.4 & 63.4 \\
    \textbf{Ours} & 57.2 & 82.2 & \textbf{64.0} \\
    \Xhline{2.8\arrayrulewidth}
\end{tabular}
\end{threeparttable}
\end{center}
\end{table}

\begin{table}[t]
\small
\begin{center}
\caption{Comparison of transferring performance on iNaturalist fine-grained classification.}
\label{tab:inaturalist}
\begin{threeparttable}
\begin{tabular}{
    >{\raggedright\arraybackslash} m{3.3cm}
    >{\centering\arraybackslash} m{1.0cm}
    >{\centering\arraybackslash} m{1.0cm}}
    \Xhline{2.8\arrayrulewidth}
    \textbf{Method} & \textbf{Top-1} & \textbf{Top-5} \\
    \hline
    \textit{Scratch} & 65.4 & 85.5 \\
    \textit{Supervised} & 66.0 & 85.6 \\
    MoCo~\cite{moco2020} & 65.7 & 85.7 \\
    PIC~\cite{parametric2020} & \textbf{66.2} & 85.7 \\
    \textbf{Ours} & \textbf{66.2} & \textbf{86.2} \\
    \Xhline{2.8\arrayrulewidth}
\end{tabular}
\end{threeparttable}
\end{center}
\end{table}

\begin{table}[t]
\small
\begin{center}
\caption{Transferring performance of different pretrained models on more downstream visual tasks.}
\label{tab:more_tasks}
\begin{threeparttable}
\begin{tabular}{
    >{\raggedright\arraybackslash} m{1.8cm}
    >{\centering\arraybackslash} m{1.2cm}
    >{\centering\arraybackslash} m{1.4cm}
    >{\centering\arraybackslash} m{1.1cm}
    >{\centering\arraybackslash} m{0.8cm}}
    \Xhline{2.8\arrayrulewidth}
    \textbf{Method} & \textbf{CIFAR10} & \textbf{CIFAR100} & \textbf{SUN397} & \textbf{DTD} \\
    \hline
    \textit{Scratch} & 95.9 & 80.2 & 53.6 & 64.8 \\
    \textit{Supervised} & 97.5 & \textbf{86.4} & \textbf{64.3} & 74.6 \\
    SimCLR & 97.7 & 85.9 & 63.5 & 73.2 \\
    \textbf{Ours} & \textbf{97.8} & \textbf{86.2} & \textbf{64.2} & \textbf{77.6} \\
    \Xhline{2.8\arrayrulewidth}
\end{tabular}
\end{threeparttable}
\end{center}
\end{table}

\subsection{Comparison with Previous Results}

% We compare representations learned by our work and by state-of-the-art unsupervised methods.
% The representation quality is measured in several ways, including ImageNet-1K linear evaluation, semi-supervised classification, and transferring experiments on multiple downstream visual tasks.

\noindent \textbf{ImageNet linear evaluation.}
Table~\ref{tab:linear_evaluation} compares our work with previous unsupervised visual representation learning methods under the ImageNet linear evaluation protocol.
We follow recent practices~\cite{simclr2020,mocov2_2020} to train a longer length, i.e., 1000 epochs.
The proposed unsupervised learning framework achieves a top-1 accuracy of 71.4\% on ImageNet-1K, outperforming SimCLR (+2.1\%), PIC (0.6\%) and MoCoV2 (+0.3\%).
The results verify that a simple full instance classification framework can learn very competitive visual representations.
The performance gains can partly be attributed to the ability of large-scale full negative instance exploration, which is not supported by previous unsupervised frameworks~\cite{simclr2020,mocov2_2020,parametric2020}.

\noindent \textbf{Semi-supervised learning.}
Following~\cite{revisiting2019,simclr2020}, we sample a 1\% or 10\% fraction of labeled data from ImageNet, and train a classifier starting from our unsupervised pretrained model to evaluate the semi-supervised learning performance.
For 1\% labels, we train the backbone with a \textit{lr} of 0.001 and the classifier with a \textit{lr} of 15.
For 10\% labels, the \textit{lr}s for the backbone and the classifier are set to 0.001 and 10, respectively~\cite{pcl2020}.
% We adopt different learning rate for the pretrained backbone and the linear classifier following~\cite{pcl2020}. For 1\% labels, we train the model for 10 epochs with learning rate (lr) of 0.001 for the backone and lr of 15 for the classifier. For 10\% labels, we train the model for 20 epochs with lr of 0.001 for the backbone and lr of 10 for the classifier.
Table~\ref{tab:semi_supervised} compares our work with both \textit{representation learning} based and \textit{label propagation} based methods.
We obtain a top-5 accuracy of 81.8\% when only 1\% of labels are used, outperforming all previous methods by a non-negligible margin (\textbf{+4.7\%}).
We also achieve the best result when 10\% of labels are provided, surpassing SimCLR and PIC by 1.4\% and 0.5\%, respectively.
The results suggest the strong discriminative ability of our learned representation.

\noindent \textbf{Transfer learning.}
To further evaluate the learned representation, we apply the pretrained model to several downstream visual tasks (including detection, fine-grained classification, and many others) to evaluate the transferring performance.

\textit{PASCAL VOC Object Detection:}
Following~\cite{moco2020}, we use Faster-RCNN~\cite{fasterrcnn2015} with ResNet-50 backbone as the object detector.
We initialize ResNet-50 with our pretrained weights, and finetune all layers end-to-end on the \textit{trainval07+12} set of the PASCAL VOC dataset~\cite{pascalvoc2010}.
We adopt the same experiment settings as MoCoV2~\cite{mocov2_2020}.
The AP (Average Precision), AP50, and AP75 scores on the \textit{test2007} set are used as indicators.
Table~\ref{tab:pascal_voc} shows the results.
Our transferring performance is significantly better than supervised pretraining counterpart (+3.7\% in AP), and is competitive with state-of-the-art unsupervised learning methods.

\textit{iNaturalist fine-grained classification:}
We finetune the pretrained model end-to-end on the \textit{train} set of iNaturalist 2018 dataset~\cite{inaturalist2018} and evaluate the top-1 and top-5 classification accuracies on the \textit{val} set.
Results are shown in Table~\ref{tab:inaturalist}.
Our method is closely competitive with the ImageNet supervised pretraining counterpart as well as previous state-of-the-art unsupervised methods.
The results indicate the discriminative ability of our pretrained representation in fine-grained classification.

\textit{More downstream tasks:}
Table~\ref{tab:more_tasks} shows transferring results on more downstream tasks, including image classification on CIFAR10, CIFAR100~\cite{cifar2009}, SUN397~\cite{sun2010}, and DTD~\cite{dtd2014}. To summarize, our method performs competitively with ImageNet supervised pretraining as well as state-of-the-art unsupervised pretraining.

\section{Conclusion}

In this work, we present an unsupervised visual representation learning framework where the pretext task is to distinguish all instances in a dataset with a parametric classifier.
The task is similar to supervised semantic classification, but with a much larger number of classes (equal to the dataset size) and finer granularity.
% We present novel techniques to make the task scalable and have higher generatlization ability.
We first introduce a hybrid parallel training framework to make large-scale instance classification feasible, which significantly reduces the GPU memory overhead and speeds up training in our experiments.
Second, we propose to improve the convergence by introducing a contrastive prior to the instance classifier.
This is achieved by initializing the classification weights as raw instance features extracted by a fixed random network with running BNs.
We show in our experiments that this simple strategy clearly speeds up convergence and improves the transferring performance.
Finally, to reduce the impact of noisy negative instance pairs, we propose to smooth the labels of a few hardest classes.
Extensive experiments on ImageNet classification, semi-supervised classification, and many downstream tasks show that our simple unsupervised representation learning method performs comparable or superior than state-of-the-art unsupervised methods.

\bibliographystyle{aaai21}
\bibliography{aaai21_instcls.bib}

\end{document}